\documentclass[conference]{IEEEtran}
\usepackage{times}

\usepackage[numbers]{natbib}
\usepackage{multicol}
\usepackage[bookmarks=true]{hyperref}

\usepackage{graphicx}        
\usepackage{url}
\usepackage{color}
\usepackage{benktools}
\usepackage{indentfirst}
\usepackage{amsmath,graphics}
\usepackage{amssymb} 
\usepackage{nicefrac, multirow}
\usepackage{algpseudocode}
\usepackage[linesnumbered]{algorithm2e}
\usepackage{tikz}
\usepackage{bigstrut}
\usepackage{pgfplots}
\pgfplotsset{compat=newest}
\usepackage{subfig}
\usepackage{textcomp}
\usetikzlibrary{backgrounds}
\numberwithin{equation}{section} 

\tikzstyle{every node}=[font=\small]

\usepackage{tabularx}
\usepackage{pbox}
\usepackage{ragged2e}
\usepackage{multirow}
\usepackage{floatrow}
\newfloatcommand{capbtabbox}{table}[][\FBwidth]

\newcolumntype{L}[1]{>{\RaggedRight\hspace{0pt}}p{#1}}
\newcolumntype{R}[1]{>{\RaggedLeft\hspace{0pt}}p{#1}}

\newcommand{\bR}{\mathbb{R}}

\renewcommand{\phi}{\varphi}

\newcommand{\furlp}[1]{\colorbox{blue!10}{\href{run:/home/fulong/academia/library/papers/#1.pdf}{D}}}
\newcommand{\furlb}[1]{\colorbox{blue!10}{\href{run:/home/fulong/academia/library/books/#1.pdf}{D}}}

\newcommand{\bx}{\mathbf{x}}

\newcommand{\bu}{\mathbf{u}}
\newcommand{\bc}{\mathbf{c}}
\newcommand{\bg}{\mathbf{g}}
\newcommand{\by}{\mathbf{y}}

\newcommand{\bp}{\mathbf{p}}

\newcommand{\bv}{\mathbf{v}}

\newcommand{\mG}{\mathcal{G}}

\newcommand{\mL}{\mathcal{L}}
\newcommand{\mN}{\mathcal{N}}
\newcommand{\mE}{\mathcal{E}}
\newcommand{\mO}{\mathcal{O}}

\newcommand{\mC}{\mathcal{C}}

\newcommand{\mU}{\mathcal{U}}

\newcommand{\mS}{\mathcal{S}}

\newcommand{\specialcell}[2][c]{ \begin{tabular}[#1]{@{}c@{}}#2\end{tabular}}

\newcommand{\myargmax}[1] {\underset{#1}{\text{argmax }}}
\newcommand{\myargmin}[1] {\underset{#1}{\text{argmin }}}

\usetikzlibrary{external}
\makeindex

\begin{document}

\title{Dex-Net 2.0: Deep Learning to Plan Robust \\ Grasps with Synthetic Point Clouds \\ and Analytic Grasp Metrics}


\author{\authorblockN{Jeffrey Mahler\authorrefmark{1},
Jacky Liang\authorrefmark{1},
Sherdil Niyaz\authorrefmark{1},
Michael Laskey\authorrefmark{1},
Richard Doan\authorrefmark{1},
Xinyu Liu\authorrefmark{1}, \\
Juan Aparicio Ojea\authorrefmark{2}, and
Ken Goldberg\authorrefmark{1}
\authorblockA{\authorrefmark{1}Dept. of EECS, University of California, Berkeley \\
Email: \{jmahler, jackyliang, sniyaz, laskeymd, rdoan, xinyuliu, goldberg\}@berkeley.edu}
\authorblockA{\authorrefmark{2} Siemens Corporation, Corporate Technology\\
Email: juan.aparicio@siemens.com}}
}


\date{January 08, 2017}



%

\maketitle

\begin{abstract}
To reduce data collection time for deep learning of robust robotic grasp plans, we explore training from a synthetic dataset of 6.7 million point clouds, grasps, and analytic grasp metrics generated from thousands of 3D models from Dex-Net 1.0 in randomized poses on a table.
We use the resulting dataset, Dex-Net 2.0, to train a Grasp Quality Convolutional Neural Network (GQ-CNN) model that rapidly predicts the probability of success of grasps from depth images, where grasps are specified as the planar position, angle, and depth of a gripper relative to an RGB-D sensor.
Experiments with over 1,000 trials on an ABB YuMi comparing grasp planning methods on singulated objects suggest that a GQ-CNN trained  with only synthetic data from Dex-Net 2.0 can be used to plan grasps in 0.8$s$ with a success rate of 93$\%$ on eight known objects with adversarial geometry and is 3$\times$ faster than registering point clouds to a precomputed dataset of objects and indexing grasps.
The Dex-Net 2.0 grasp planner also has the highest success rate on a dataset of 10 novel rigid objects and achieves 99$\%$ precision (one false positive out of 69 grasps classified as robust) on a dataset of 40 novel household objects, some of which are articulated or deformable.
Code, datasets, videos, and supplementary material are available at \url{http://berkeleyautomation.github.io/dex-net}.
\end{abstract}

\IEEEpeerreviewmaketitle


\section{Introduction}
\seclabel{introduction}
Reliable robotic grasping is challenging due to imprecision in sensing and actuation, which leads to uncertainty about properties such as object shape, pose, material properties, and mass.
Recent results suggest that deep neural networks trained on large datasets of human grasp labels~\cite{lenz2015deep} or physical grasp outcomes~\cite{pinto2016supersizing} can be used to plan grasps that are successful across a wide variety of objects directly from images or point clouds, similar to generalization results in computer vision~\cite{krizhevsky2012imagenet}.
However, data collection requires either tedious human labeling~\cite{kappler2015leveraging} or months of execution time on a physical system~\cite{levine2016learning}.

An alternative approach is to plan grasps using physics-based analyses such as caging~\cite{rodriguez2012caging}, grasp wrench space (GWS) analysis~\cite{prattichizzo2008grasping}, robust GWS analysis~\cite{weisz2012pose}, or simulation~\cite{ kappler2015leveraging}, which can be rapidly computed using Cloud Computing~\cite{kehoe2013cloud}.
However, these methods assume a separate perception system that estimates properties such as object shape or pose either perfectly~\cite{prattichizzo2008grasping} or according to known Gaussian distributions~\cite{mahler2016dexnet}. 
This is prone to errors~\cite{balasubramanian2012physical}, may not generalize well to new objects, and can be slow to match point clouds to known models during execution~\cite{goldfeder2011data}.
In this paper we instead consider predicting grasp success directly from depth images by training a deep Convolutional Neural Network (CNN) on a massive dataset of parallel-jaw grasps, grasp metrics, and rendered point clouds generated using analytic models of robust grasping and image formation~\cite{hartley2003multiple, mallick2014characterizations}, building upon recent research on classifying force closure grasps~\cite{gualtieri2016high, seitalarge} and the outcomes of dynamic grasping simulations~\cite{johns2016deep, kappler2015leveraging, varley2015generating}. 

\begin{figure}[t!]
\centering
\includegraphics[scale=0.2]{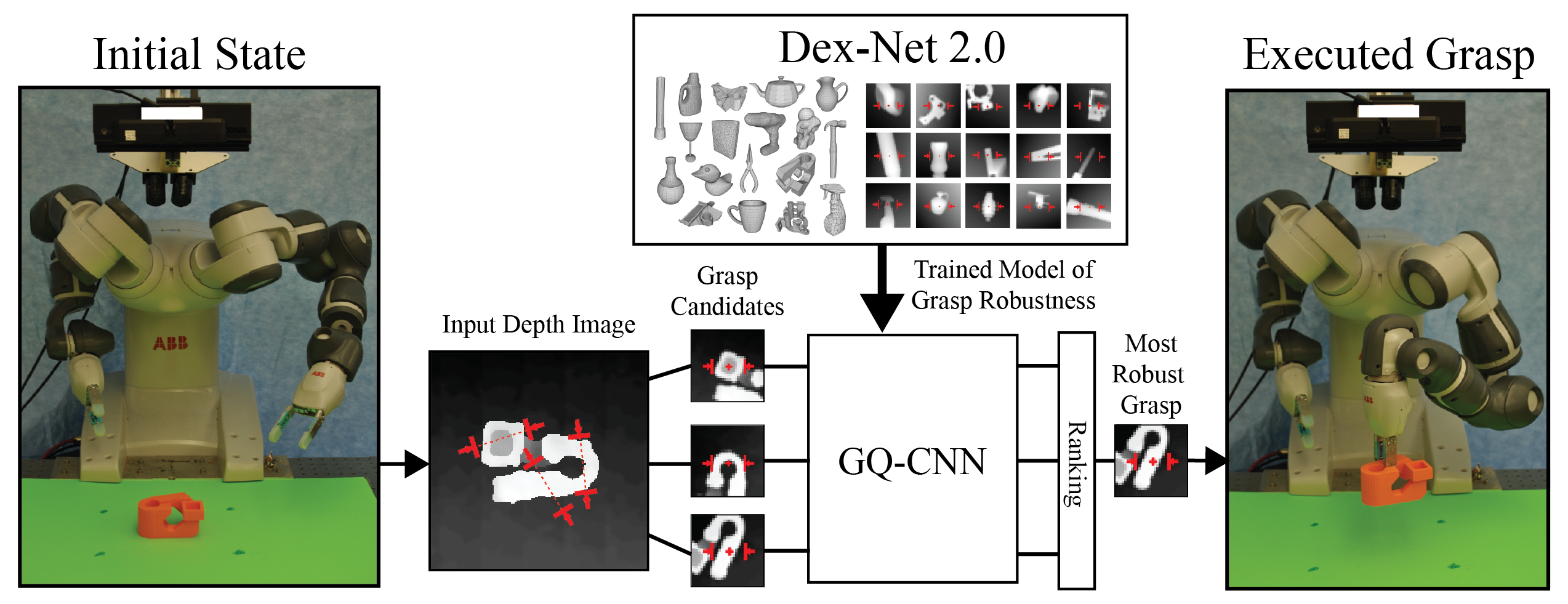}
\caption{Dex-Net 2.0 Architecture. (Center) The Grasp Quality Convolutional Neural Network (GQ-CNN) is trained offline to predict the robustness candidate grasps from depth images using a dataset of 6.7 million synthetic point clouds, grasps, and associated robust grasp metrics computed with Dex-Net 1.0. (Left) When an object is presented to the robot, a depth camera returns a 3D point cloud, where pairs of antipodal points identify a set of several hundred grasp candidates. (Right) The GQ-CNN rapidly determines the most robust grasp candidate, which is executed with the ABB YuMi robot.
\vspace*{-20pt}}
\figlabel{teaser}
\end{figure}

Our primary contributions are: 1) the Dexterity Network (Dex-Net) 2.0, a dataset associating 6.7 million point clouds and analytic grasp quality metrics with parallel-jaw grasps planned using robust quasi-static GWS analysis on a dataset of 1,500 3D object models, 2) a Grasp Quality Convolutional Neural Network (GQ-CNN) model trained to classify robust grasps in depth images using expected epsilon quality as supervision, where each grasp is specified as a planar pose and depth relative to a camera, and 3) a grasp planning method that samples antipodal grasp candidates and ranks them with a GQ-CNN.

In over 1,000 physical trials of grasping single objects on a tabletop with an ABB YuMi robot, we compare Dex-Net 2.0 to image-based grasp heuristics, a random forest~\cite{seitalarge}, an SVM~\cite{ten2015using}, and a baseline that recognizes objects, registers their 3D pose~\cite{goldfeder2011data}, and indexes Dex-Net 1.0~\cite{mahler2016dexnet} for the most robust grasp to execute.
We find that the Dex-Net 2.0 grasp planner is 3$\times$ faster than the registration-based method, 93$\%$ successful on objects seen in training (the highest of learning-based methods), and is the best performing method on novel objects, achieving 99$\%$ precision on a dataset of 40 household objects despite being trained entirely on synthetic data.

\section{Related Work}
\seclabel{related-work}

{\bf Grasp Planning.} Given an object and reachability constraints due to the environment, grasp planning considers finding a gripper configuration that maximizes a success (or quality) metric.
Methods fall into one of two categories based on success criteria: {\it analytic} methods~\cite{prattichizzo2008grasping}, which consider performance according to physical models such as the ability to resist external wrenches~\cite{pokorny2013classical}, and {\it empirical} (or data-driven) methods~\cite{bohg2014data}, which typically use human labels~\cite{balasubramanian2012physical} or the ability to lift the object in physical trials~\cite{pinto2016supersizing}.

{\it Analytic Methods.}
Analytic approaches typically assume that object and contact locations are known exactly and consider either the ability to resist external wrenches~\cite{prattichizzo2008grasping} or the ability to constrain the object's motion~\cite{rodriguez2012caging}.
To execute grasps on a physical robot, a common approach is to precompute a database of known 3D objects labeled with grasps and quality metrics such as GraspIt!~\cite{goldfeder2009columbia}.
Precomputed grasps are indexed using {\it point cloud registration}: matching point clouds to known 3D object models in the database using visual and geometric similarity~\cite{bohg2014data, brook2011collaborative, ciocarlie2014towards, goldfeder2011data, hernandez2016team, hinterstoisser2011multimodal, kehoe2013cloud} and executing the highest quality grasp for the estimated object instance.

Robust grasp planning (RGP) methods maximize grasp {\it robustness}, or the expected value of an analytic metric under uncertainty in sensing and control.
This typically involves labeling grasps on a database 3D object models with robust metrics such as probability of force closure ~\cite{kehoe2013cloud} or the pose error robust metric~\cite{weisz2012pose} and using registration-based planning.
Recent research has demonstrated that the sampling complexity of RGP can be improved using Multi-Armed Bandits~\cite{laskey2015bandits} and datasets of prior 3D objects and robust grasps, such as the Dexterity Network (Dex-Net) 1.0~\cite{mahler2016dexnet}.
In this work we make a major extension to Dex-Net 1.0 by associating synthetic point clouds with robust grasps and training a Convolutional Neural Network to map point clouds and candidates grasps to estimated robustness.

{\it Empirical Methods.}
Empirical approaches typically use machine learning to develop models that map from robotic sensor readings directly to success labels from humans or physical trials.
Human labels have become popular due to empirical correlation with physical success~\cite{balasubramanian2012physical}, although they may be expensive to aquire for large datasets.
Research in this area has largely focused on associating human labels with graspable regions in RGB-D images~\cite{lenz2015deep} or point clouds~\cite{detry2013learning, herzog2014learning, kappler2015leveraging}.
Lenz et al.~\cite{lenz2015deep} created a dataset of over 1k RGB-D images with human labels of successful and unsuccessful grasping regions, which has been used to train fast CNN-based detection models~\cite{redmon2015real}.

Another line of research on empirical grasp planning has attempted to optimize success in physical trials directly.
The time cost of generating samples on a physical robot led to the development active methods for acquiring grasping experiences such as Multi-Armed Bandits using Correlated Beta Processes~\cite{montesano2012active} or Prior Confidence Bounds~\cite{oberlin2015autonomously}.
Recently Pinto and Gupta~\cite{pinto2016supersizing} scaled up data collection by recording over 40k grasping experiences on a Baxter and iteratively training CNNs to predict lifting successes or to resist grasp perturbations caused by an adversary~\cite{pinto2016supervision}.
Levine et al.~\cite{levine2016learning} scaled up dataset collection even futher, collecting over 800k datapoints with a set of continuously running robotic arms and using deep learning to predict end effector poses.
However, this required over 2 months of training across up to 14 robots.

{\bf Computer Vision Techniques in Robot Grasping.}
Grasps for a physical robot are typically planned from images of target objects. 
Analytic grasp planning methods register images of rigid objects to a known database of 3D models, which typically involves segmentation, classification, and geometric pose estimation~\cite{ciocarlie2014towards, hinterstoisser2011multimodal, salas2013slam++, xie2013multimodal} from 3D point cloud data in order to index precomputed grasps.
However, this multi-stage approach can be prone to many hyperparameters that are difficult to tune and errors can compound across modules.

An alternative approach is to use deep learning to estimate 3D object shape and pose directly from color and depth images~\cite{gupta2015aligning, zeng2016multi}.
Recent research in robotics has focused on how to improve accuracy in object recognition by structuring the way the neural network fuses the separate color and depth streams from images~\cite{porzi2016learning} and adding synthetic noise to synthetic training images~\cite{eitel2015multimodal}.
Another approach is to detect graspable regions directly in images without explicitly representing object shape and pose~\cite{lenz2015deep, nguyen2016detecting, pinto2016supersizing, saxena2008robotic}, as it may not always be necessary to explicitly recognize objects and their pose to perform a grasp.

Since training on real images may require significant data collection time, an alternative approach is to learn on simulated images~\cite{varley2015generating, veres2017modeling} and to adapt the representation to real data~\cite{tzeng2016adapting, zhu2016target}.
Recent research suggests that in some cases it may be sufficient to train on datasets generated using perturbations to the parameters of the simulator~\cite{gualtieri2016high, sadeghi2016cad}.
Notably, Johns et al.~\cite{johns2016deep} used rendered depth images with simulated noise to train a CNN-based detector for planar grasps using dynamic simulations as supervision.
We build upon these results by using robust analytic grasp metrics as supervision, using the gripper's distance from the camera in predictions, and performing extensive evaluations on a physical robot.

\section{Problem Statement}
\seclabel{problem-statement}

We consider the problem of planning a robust planar parallel-jaw grasp for a singulated rigid object resting on a table based on point clouds from a depth camera.
We learn a function that takes as input a candidate grasp and a depth image and outputs an estimate of robustness~\cite{kehoe2013cloud, weisz2012pose}, or probability of success under uncertainty in sensing and control.

\subsection{Assumptions}
\seclabel{assumptions}
We assume a parallel-jaw gripper, rigid objects singulated on a planar worksurface, and single-view (2.5D) point clouds taken with a depth camera.
For generating datasets, we assume a known gripper geometry and a single overhead depth camera with known intrinsics.

\subsection{Definitions}
\seclabel{definitions}

\begin{figure}[t!]
\centering
\includegraphics[scale=0.4]{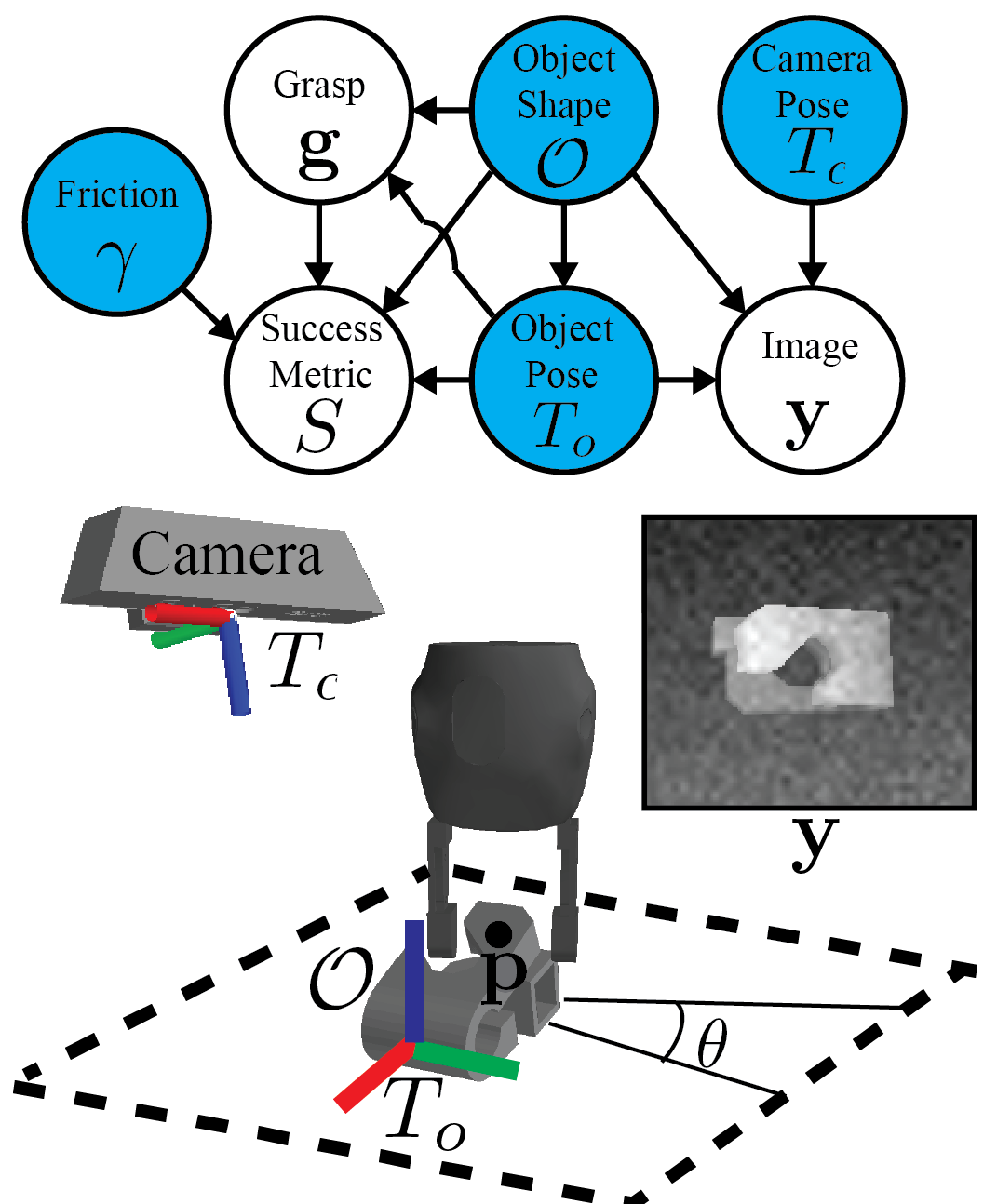}
\caption{Graphical model for robust parallel-jaw grasping of objects on a table surface based on point clouds. Blue nodes are variables included in the state representation. Object shapes $\mO$ are uniformly distributed over a discrete set of object models and object poses $T_{o}$ are distributed over the object's stable poses and a bounded region of a planar surface. Grasps $\bu = (\bp, \phi)$ are sampled uniformly from the object surface using antipodality constraints. Given the coefficient of friction $\gamma$ we evaluate an analytic success metric $S$ for a grasp on an object. A synthetic 2.5D point cloud $\by$ is generated from 3D meshes based on the camera pose $T_{c}$, object shape, and pose and corrupted with multiplicative and Gaussian Process noise.
\vspace*{-20pt}}
\figlabel{graphical-model}
\end{figure}


{\bf States.} Let $\bx = (\mO, T_{o}, T_{c}, \gamma)$ denote a state describing the variable properties of the camera and objects in the environment, where $\mO$ specifies the geometry and mass properties of an object, $T_{o}, T_{c}$ are the 3D poses of the object and camera, respectively, and $\gamma \in \bR$ is the coefficient of friction between the object and gripper.

{\bf Grasps.} Let $\bu = (\bp, \phi) \in \bR^3 \times \mS^1$ denote a parallel-jaw grasp in 3D space specified by a center $\bp = (x, y, z) \in \bR^3$ relative to the camera and an angle in the table plane $\phi \in \mS^1$.

{\bf Point Clouds.} Let $\by = \bR_{+}^{H \times W}$ be a 2.5D point cloud represented as a depth image with height $H$ and width $W$ taken by a camera with known intrinsics~\cite{hartley2003multiple}, and let $T_{c}$ be the 3D pose of the camera.

{\bf Robust Analytic Grasp Metircs.} 
Let $S(\bu, \bx) \in \{0, 1\}$ be a binary-valued grasp success metric, such as force closure or physical lifting.
Let $p(S, \bu, \bx, \by)$ be a joint distribution on grasp success, grasps, states, and point clouds modeling imprecision in sensing and control.
For example, $p$ could be defined by noisy sensor readings of a known set of industrial parts coming down a conveyor belt in arbitrary poses.
Let the {\it robustness} of a grasp given an observation~\cite{brook2011collaborative, weisz2012pose} be the expected value of the metric, or probability of success under uncertainty in sensing and control: $Q(\bu, \by) = \mathbb{E} \left[ S \mid \bu, \by \right].$

\begin{figure*}[t!]
\centering
\includegraphics[scale=0.4]{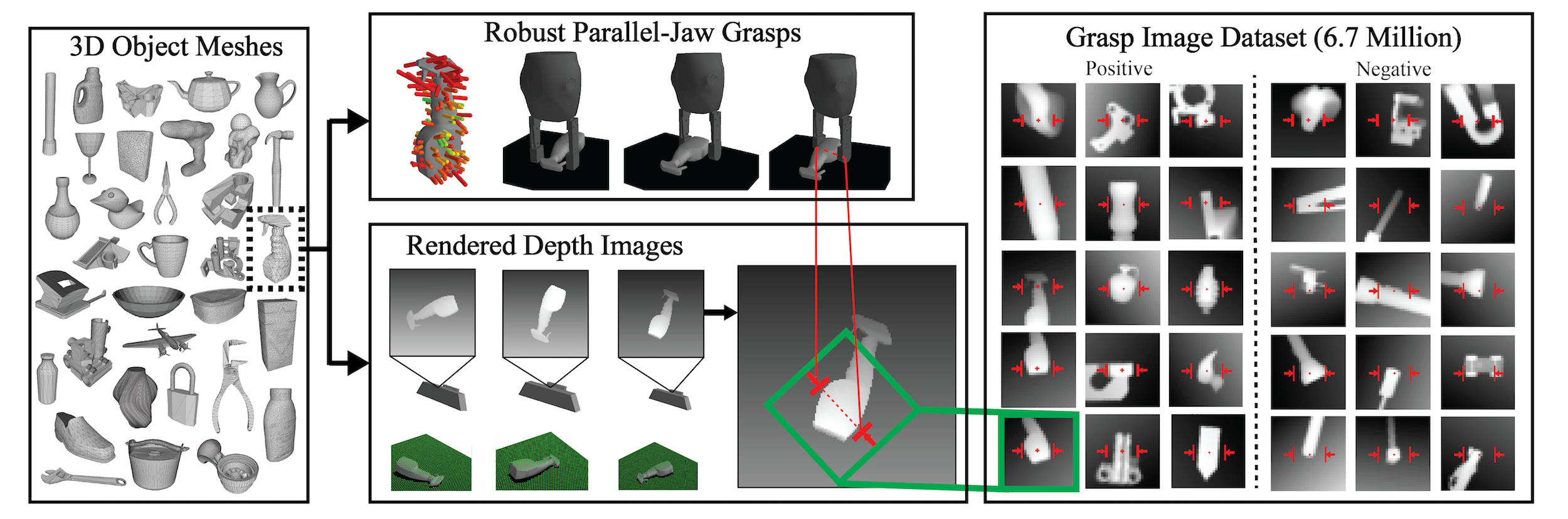}
\caption{Dex-Net 2.0 pipeline for training dataset generation. (Left) The database contains 1,500 3D object mesh models. (Top) For each object, we sample hundreds of parallel-jaw grasps to cover the surface and evaluate robust analytic grasp metrics using sampling. For each stable pose of the object we associate a set of grasps that are perpendicular to the table and collision-free for a given gripper model. (Bottom) We also render point clouds of each object in each stable pose, with the planar object pose and camera pose sampled uniformly at random. Every grasp for a given stable pose is associated with a pixel location and orientation in the rendered image. (Right) Each image is rotated, translated, cropped, and scaled to align the grasp pixel location with the image center and the grasp axis with the middle row of the image, creating a $32\times32$ grasp image. The full dataset contains over 6.7 million grasp images.}
\figlabel{dataset}
\end{figure*}

\subsection{Objective}
\seclabel{objective}
Our goal is to learn a robustness function $Q_{\theta^*}(\bu, \by) \in [0,1]$ over many possible grasps, objects, and images that classifies grasps according to the binary success metric:
\begin{align}
	\theta^* = \myargmin{\theta \in \Theta} \mathbb{E}_{p(S,\bu,\bx,\by)}\left[ \mL(S, Q_{\theta}(\bu, \by)) \right]  \label{eq:obj}
\end{align}
\noindent where $\mL$ is the cross-entropy loss function and $\Theta$ defines the parameters of the Grasp Quality Convolutional Network (GQ-CNN) described in \secref{GQ-CNN}.
This objective is motivated by that fact that $Q_{\theta^*} = Q$ for all possible grasps and images when there exists some $\theta \in \Theta$ such that $Q_{\theta} = Q$~\cite{miller1993loss}.
The estimated robustness function can be used in a grasping policy that maximizes $Q_{\theta^*}$ over a set of candidate grasps: $\pi_{\theta}(\by) = \text{argmax}_{\bu \in \mC} Q_{\theta}(\bu, \by)$, where $\mC$ specifies constraints on the set of available grasps, such as collisions or kinematic feasibility.
Learning $Q$ rather than directly learning the policy allows us to enforce task-specific constraints without having to update the learned model. 
\section{Learning a Grasp Robustness Function}
\seclabel{methodology}

Solving for the grasp robustness function in objective~\ref{eq:obj} is challenging for several reasons.
First, we may need a huge number of samples to approximate the expectation over a large number of possible objects.
We address this by generating Dex-Net 2.0, a training dataset of 6.7 million synthetic point clouds, parallel-jaw grasps, and robust analytic grasp metrics across 1,500 3D models sampled from the graphical model in \figref{graphical-model}. 
Second, the relationship between point clouds, grasps, and metrics over a large datset of objects may be complex and difficult to learn with linear or kernelized models.
Consequently, we develop a Grasp Quality Convolutional Neural Network (GQ-CNN) model that classifies robust grasp poses in depth images and train the model on data from Dex-Net 2.0.

\subsection{Dataset Generation}
\seclabel{dataset}

We estimate $Q_{\theta^*}$ using a sample approximation~\cite{friedman2001elements} of the objective in Equation~\ref{eq:obj} using i.i.d samples $(S_1, \bu_1, \bx_1, \by_1), ..., (S_N, \bu_N, \bx_N, \by_N) \sim p(S, \bu, \bx, \by)$ from our generative graphical model for images, grasps, and success metrics:
\vspace{-1ex}
\begin{align*}
	\hat{\theta} = \myargmin{\theta \in \Theta} \sum \limits_{i=1}^N \mL(S_i, Q_{\theta}(\bu_i, \by_i)).
\end{align*}

\subsubsection{Graphical Model}
\seclabel{graphical}
Our graphical model is illustrated in \figref{graphical-model} and models $p(S, \bu, \bx, \by)$ as the product of a state distribution $p(\bx)$, an observation model $p(\by | \bx)$, a grasp candidate model $p(\bu | \bx)$, and an analytic model of grasp success $p(S | \bu, \bx)$.

\begin{table}[t]
\centering
\resizebox{\textwidth}{!}{%
   \begin{tabular}{| r | c |}
   \hline
  {\bf Distribution} & {\bf Description}\\ \hline
  $p(\gamma)$ & truncated Gaussian distribution over friction coefficients \\
  \hline
  $p(\mO)$ & discrete uniform distribution over 3D object models \\
  \hline
  $p(T_{o} | \mO)$ & \begin{tabular}[x]{@{}c@{}} continuous uniform distribution over the discrete set of \\ object stable poses and planar poses on the table surface \end{tabular}  \\
  \hline
  $p(T_{c})$ & \begin{tabular}[x]{@{}c@{}} continuous uniform distribution over spherical coordinates \\ for radial bounds $[r_{\ell}, r_{u}]$ and polar angle in $[0, \delta]$ \end{tabular} \\
  \hline
  \end{tabular}}
        \caption{Details of the distributions used in the Dex-Net 2.0 graphical model for generating the Dex-Net training dataset.  }
		\tablabel{distributions}
\end{table}

We model the state distribution as
\begin{align*}
	p(\bx) &= p(\gamma) p(\mO) p(T_{o} | \mO)  p(T_{c})
\end{align*}
\noindent where the distributions are detailed in \tabref{distributions}.
Our grasp candidate model $p(\bu \mid \bx)$ is a uniform distribution over pairs of antipodal contact points on the object surface that form a grasp axis parallel to the table plane.
Our observation model is $\by = \alpha \hat{\by} + \epsilon$ where $\hat{\by}$ is a rendered depth image for a given object in a given pose, $\alpha$ is a Gamma random variable modeling depth-proportional noise, and $\epsilon$ is zero-mean Gaussian Process noise over pixel coordinates with bandwidth $\ell$ and measurement noise $\sigma$ modeling additive noise~\cite{mallick2014characterizations}.
We model grasp success as: 
\begin{align*}
	S(\bu, \bx) = \left\{ \begin{array}{cc} 1 & E_{Q} > \delta \text{ and } collfree(\bu, \bx) \\ 0 & otherwise \end{array} \right.
\end{align*}
\noindent where $E_{Q}$ is the robust epsilon quality defined in~\cite{seitalarge}, a variant of the pose error robust metric~\cite{weisz2012pose} that includes uncertainty in friction and gripper pose, and $collfree(\bu, \bx)$ indicates that the gripper does not collide with the object or table.
The supplemental file details the parameters of these distributions.

\begin{figure*}[t!]
\centering
\includegraphics[scale=0.26]{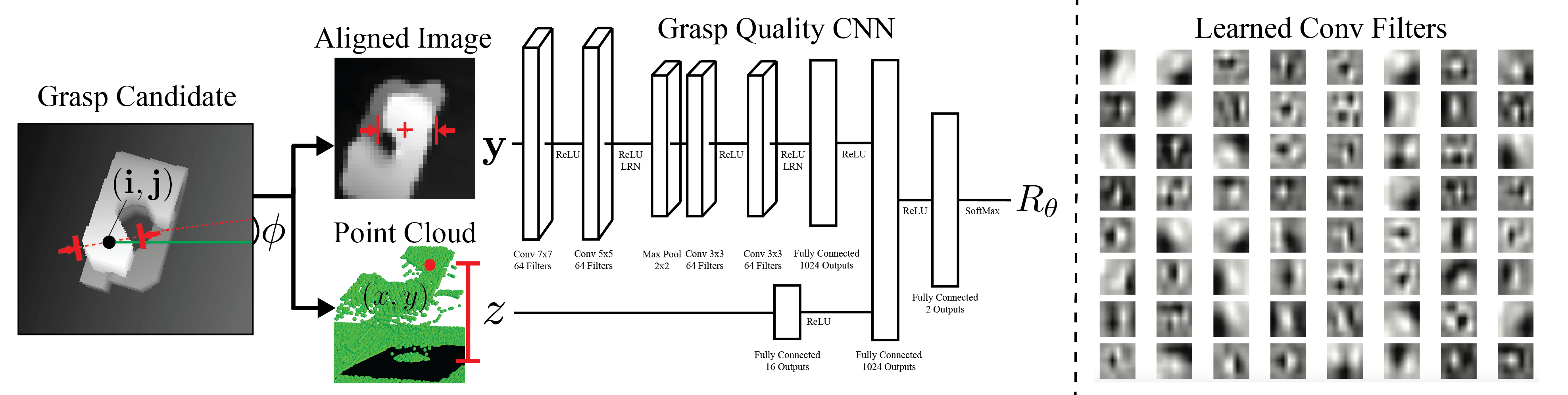}
\caption{(Left) Architecture of the Grasp Quality Convolutional Neural Network (GQ-CNN). Planar grasp candidates $\bu = (i,j,\phi,z)$ are generated from a depth image and transformed to align the image with the grasp center pixel $(i,j)$ and orientation $\phi$. The architecture contains four convolutional layers in pairs of two separated by ReLU nonlinearities followed by 3 fully connected layers and a separate input layer for the $z$, the distance of the gripper from the camera. The use of convolutional layers was motivated by the relevance of depth edges as features for learning in previous research~\cite{bohg2010learning, lenz2015deep, mahler2016dexnet} and the use of ReLUs was motivated by image classification results~\cite{krizhevsky2012imagenet}. The network estimates the probability of grasp success (robustness) $Q_{\theta} \in [0,1]$, which can be used to rank grasp candidates. (Right) The first layer of convolutional filters learned by the GQ-CNN on Dex-Net 2.0. The filters appear to compute oriented image gradients at various scales, which may be useful for inferring contact normals and collisions between the gripper and object.}
\figlabel{architecture}
\end{figure*}

\subsubsection{Database}
\seclabel{database}
Dex-Net 2.0 contains 6.7 million datapoints generated using the pipeline of \figref{dataset}.

{\it 3D Models.} The dataset contains a subset of 1,500 mesh models from Dex-Net 1.0: 1,371 synthetic models from 3DNet~\cite{wohlkinger20123dnet} and 129 laser scans from the KIT object database~\cite{kasper2012kit}.
Each mesh is aligned to a standard frame of reference using the principal axes, rescaled to fit within a gripper width of $5.0cm$ (the opening width of an ABB YuMi gripper), and assigned a mass of $1.0kg$ centered in the object bounding box since some meshes are nonclosed.
For each object we also compute a set of stable poses~\cite{goldberg1999part} and store all stable poses with probability of occurence above a threshold.

{\it Parallel-Jaw Grasps.} Each object is labeled with a set of up to 100 parallel-jaw grasps.
The grasps are sampled using the rejection sampling method for antipodal point pairs developed in Dex-Net 1.0~\cite{mahler2016dexnet} with constraints to ensure coverage of the object surface~\cite{mahler2016privacy}.
For each grasp we evaluate the expected epsilon quality $E_Q$~\cite{pokorny2013classical} under object pose, gripper pose, and friction coefficient uncertainty using Monte-Carlo sampling~\cite{seitalarge}, 

{\it Rendered Point Clouds.} Every object is also paired with a set of 2.5D point clouds (depth images) for each object stable pose, with camera poses and planar object poses sampled according to the graphical model described in \secref{graphical}.
Images are rendered using a pinhole camera model and perspective projection with known camera intrinsics, and each rendered image is centered on the object of interest using pixel transformations.
Noise is added to the images during training as described in \secref{optimization}.

\subsection{Grasp Quality Convolutional Neural Network}
\seclabel{GQ-CNN}

\subsubsection{Architecture}
\seclabel{architecture}
The Grasp Quality Convolutional Neural Network (GQ-CNN) architecture, illustrated in \figref{architecture} and detailed in the caption, defines the set of parameters $\Theta$ used to represent the grasp robustness function $Q_{\theta}$.
The GQ-CNN takes as input the gripper depth from the camera $z$ and a depth image centered on the grasp center pixel $\bv = (i,j)$ and aligned with the grasp axis orientation $\phi$.
The image-gripper alignment removes the need to learn rotational invariances that can be modeled by known, computationally-efficient image transformations (similar to spatial transformer networks~\cite{jaderberg2015spatial}) and allows the network to evaluate any grasp orientation in the image rather than a predefined discrete set as in~\cite{johns2016deep, pinto2016supersizing}.
Following standard preprocessing conventions, we normalize the input data by subtracting the mean and dividing by the standard deviation of the training data and then pass the image and gripper depth through the network to estimate grasp robustness.
The GQ-CNN has approximately 18 million parameters.

\subsubsection{Training Dataset}
\seclabel{training-dataset}
GQ-CNN training datasets are generated by associating grasps with a pixel $\bv$, orientation $\phi$, and depth $z$ relative to rendered depth images as illustrated in \figref{dataset}.
We compute these parameters by transforming grasps into the camera frame of reference using the camera pose $T_{c}$ and projecting the 3D grasp position and orientation onto the imaging plane of the camera~\cite{hartley2003multiple}.
We then transform all pairs of images and grasp configurations to a single image centered on $\bv$ and oriented along $\phi$ (see the left panel of \figref{architecture} for an illustration).
The Dex-Net 2.0 training dataset contains 6.7 million datapoints and approximately $21.2\%$ positive examples for the thresholded robust epsilon quality with threshold $\delta = 0.002$~\cite{kappler2015leveraging} and a custom YuMi gripper.

\subsubsection{Optimization}
\seclabel{optimization}
We optimize the parameters of the GQ-CNN using backpropagation with stochastic gradient descent and momentum~\cite{krizhevsky2012imagenet}.
We initialize the weights of the model by sampling from a zero mean gaussian with variance $\frac{2}{n_i}$, where $n_i$ is the number of inputs to the $i$-th network layer~\cite{he2015delving}.
To augment the dataset, we reflect the image about its vertical and horizontal axes and rotate each image by $180^{\circ}$ since these lead to equivalent grasps.
We also adaptively sample image noise from our noise model (see \secref{graphical}) before computing the batch gradient for new samples during training to model imaging noise without explicitly storing multiple versions of each image.
To speed up noise sampling we approximate the Gaussian Process noise by upsampling an array of uncorrelated zero-mean Gaussian noise using bilinear interpolation. 
We set hyperparameters based on the performance on a randomize synthetic validation set as described in \secref{baselines}.

\section{Grasp Planning}
\seclabel{planning}
The Dex-Net 2.0 grasp planner uses the robust grasping policy $\pi_{\theta}(\by) = \text{argmax}_{\bu \in \mC} Q_{\theta}(\bu, \by)$ illustrated in \figref{teaser}.
The set $\mC$ is a discrete set of antipodal candidate grasps~\cite{chen1993finding} sampled uniformly at random in image space for surface normals defined by the depth image gradients.
Each grasp candidate is evaluated by the GQ-CNN, and the most robust grasp that is (a) kinematically reachable and (b) not in collision with the table is executed.
The supplemental file contains additional details.
\section{Experiments}
\seclabel{experiments}

We evaluated clasification performance on both real and synthetic data and performed extensive physical evaluations on an ABB YuMi with custom silicone gripper tips designed by Guo et al.~\cite{guo2017design} to benchmark the performance of grasping a single object.
All experiments ran on a Desktop running Ubuntu 14.04 with a 2.7 GHz Intel Core i5-6400 Quad-Core CPU and an NVIDIA GeForce 980, and we used an NVIDIA GeForce GTX 1080 for training large models.

\begin{figure}[t!]
\centering
\includegraphics[scale=0.4]{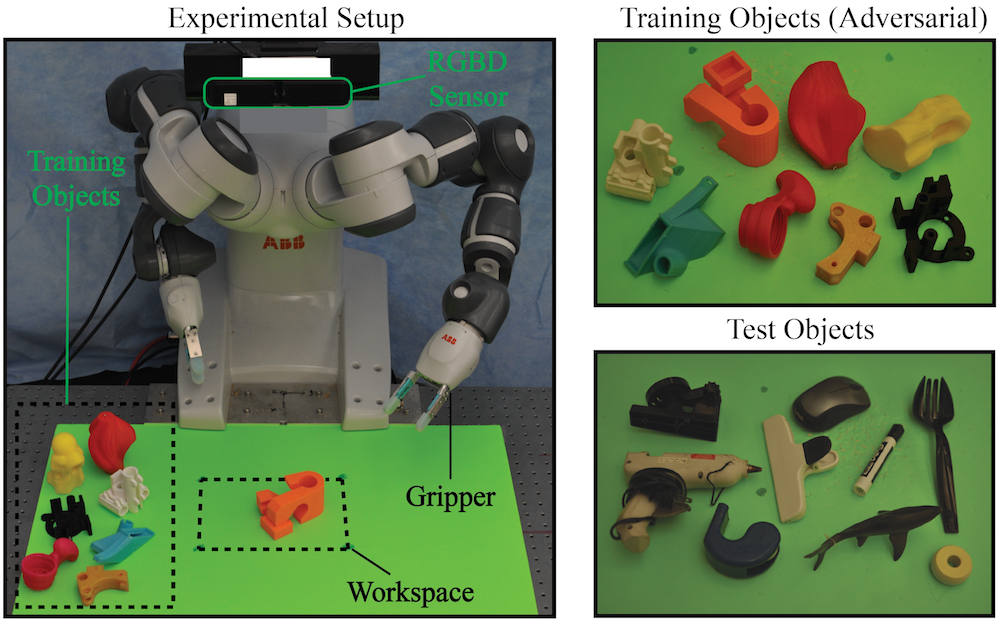}
\caption{(Left) The experimental platform for benchmarking grasping with the ABB YuMi. We registered the camera to the robot with a chessboard before each experiment. In each trial a human operator sampled an object pose by shaking the object in a box and placing it upside down in the workspace. We then took an RGB-D image with a Primsense Carmine 1.08, filled in the image using inpainting~\cite{johns2016deep}, segmented the object using color background subtraction, and formed a bounding box for the detected object. The grasp planner under evaluation then planned a gripper pose and the YuMi executed the grasp. Grasps were considered successful if the gripper held the object after lifting, transporting, and shaking the object. (Top-Right) The training set of 8 objects with adversarial geometric features such as smooth curved surfaces and narrow openings for grasping known objects. (Bottom-Right) The test set of 10 household objects not seen during training.
\vspace*{-20pt}}
\figlabel{experimental-setup}
\end{figure}

\subsection{Physical Benchmark Description}
\seclabel{benchmark}
We created a benchmark for grasping single objects on a tabletop to compare grasp planning methods.
The setup is illustrated in \figref{experimental-setup} and the experimental procedure is described in the caption and shown in the supplemental video\footnote{https://youtu.be/9eqAxk95I3Y}.
Each grasp planner received as input a color image, depth image, bounding box containing the object, and camera intrinsics, and output a target grasping pose for the gripper.
A human operator was required to reset the object in the workspace on each trial, and therefore blinded operators from which grasp planning method was being tested in order to remove bias.

We compared performance on this benchmark with the following metrics:
\begin{enumerate}
	\item {\bf Success Rate:} The percentage of grasps that were able to lift, transport, and hold a desired object after shaking.
	\item {\bf Precision:} The success rate on grasps that are have an estimated robustness higher than 50$\%$. This measures performance when the robot can decide not to grasp an object, which could be useful when the robot has other actions (e.g. pushing) available.
	\item {\bf Robust Grasp Rate:} The percentage of planned grasps with an estimated robustness higher than 50$\%$.
	\item {\bf Planning Time:} The time in seconds between receiving an image and returning a planned grasp.
\end{enumerate} 

\subsection{Datasets}
\figref{experimental-setup} illustrates the physical object datasets used in the benchmark:
\begin{enumerate}
	\item {\bf Train:} A validation set of 8 3D-printed objects with adversarial geometric features such as smooth, curved surfaces. This is used to set model parameters and to evaluate performance on known objects.  
	\item {\bf Test:} A set of 10 household objects similar to models in Dex-Net 2.0 with various material, geometric, and specular properties. This is used to evaluate generalization to unknown objects.
\end{enumerate}
\noindent We chose objects based on geometric features under three constraints: (a) small enough to fit within the workspace, (b) weight less than 0.25$kg$, the payload of the YuMi, and (c) height from the table greater than 1.0$cm$ due to a limitation of the silicone gripper fingertips.

We used four different GQ-CNN training datasets to study the effect on performance, each with a 80-20 image-wise training and validation split:
\begin{enumerate}
	\item {\bf Adv-Synth:} Synthetic images and grasps for the adversarial objects in Train (189k datapoints).
	\item {\bf Adv-Phys:} Outcomes of executing random antipodal grasps with random gripper height and friction coefficient of $\mu=0.5$ in 50 physical trials per object in Train (400 datapoints).
	\item {\bf Dex-Net-Small:} A subset of data from 150 models sampled uniformly from Dex-Net 2.0 (670k datapoints).
	\item {\bf Dex-Net-Large:} Data from all 1500 models in Dex-Net 2.0 (6.7m datapoints).
\end{enumerate}


\begin{figure}
\begin{floatrow}
\resizebox{0.5\textwidth}{!}{%
\ffigbox{%
\includegraphics[scale=0.28]{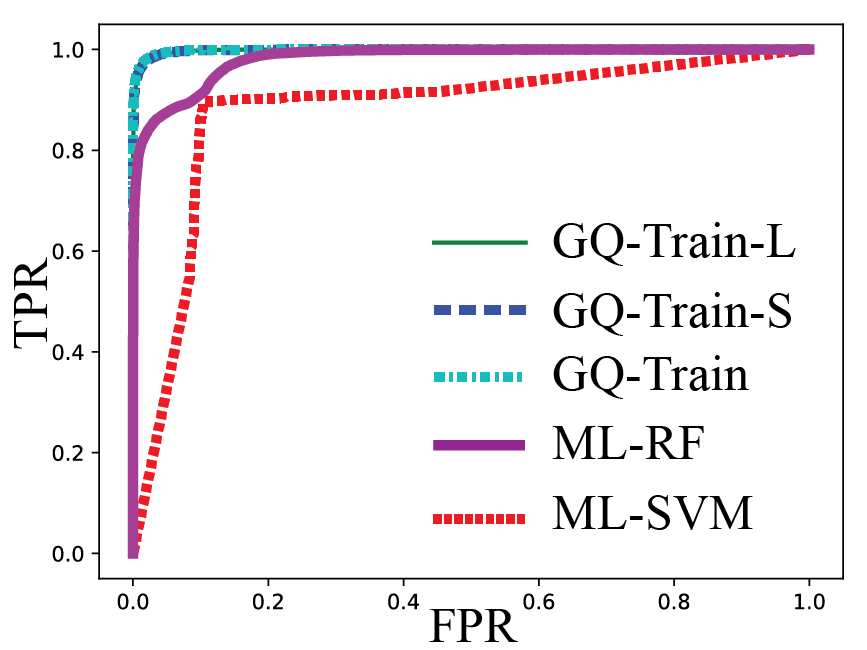}
}{
  \caption{\footnotesize Receiver operating characteristic comparing the performance of learning models on Adv-Synth. The GQ-CNN models all perform similarly and have a significantly higher true positive rate when compared to ML-RF and ML-SVM.}%
  \figlabel{roc}
}
}
\resizebox{0.5\textwidth}{!}{%
\capbtabbox{%
  \footnotesize
  \begin{tabular}{r|c}
  {\bf Model} & {\bf Accuracy ($\%$)}\\ \hline
  ML-SVM &  89.7 \\
  ML-RF &  90.5 \\
  GQ-S-Adv &  97.8 \\
  GQ-L-Adv &  97.8 \\
  GQ-Adv &  98.1 \\
  & \\
  \end{tabular}
}{\footnotesize
  \caption{The classification accuracy of each model on Adv-Synth. We see that the GQ-CNN methods have less than 2.5$\%$ test error while ML-RF and ML-SVM are closer to 10$\%$ error. Pretraining does not appear to affect performance.}
  \tablabel{synth-val}
}
}
\end{floatrow}
\end{figure}

\begin{table*}[t]
\centering
\resizebox{\textwidth}{!}{%
        \begin{tabular}{r || R{1.0cm} | R{1.0cm} | R{1.0cm} | R{1.0cm} | R{1.0cm} | R{1.0cm} || R{1.0cm} | R{1.0cm} | R{1.0cm} | R{1.0cm} | R{1.0cm} | R{1.0cm} }
       \multicolumn{1}{c ||}{\specialcell{}}& \multicolumn{6}{c ||}{\specialcell{\bf Comparions of Methods}} &\multicolumn{6}{c}{\specialcell{\bf GQ-CNN Parameter Sensitivity}} \\
       \hline & & & & & & & & & & & & \\
       \multicolumn{1}{c ||}{\specialcell{}}&  \multicolumn{1}{c |}{\specialcell{\bf Random}} & \multicolumn{1}{c |}{\specialcell{\bf IGQ}} & \multicolumn{1}{c |}{\specialcell{\bf ML-RF}} & \multicolumn{1}{c |}{\specialcell{\bf ML-SVM}} & \multicolumn{1}{c |}{\specialcell{\bf REG}} & \multicolumn{1}{c ||}{\specialcell{\bf GQ-L-Adv}} & \multicolumn{1}{c |}{\specialcell{\bf GQ-S-Adv}} & \multicolumn{1}{c |}{\specialcell{\bf GQ-Adv}} & \multicolumn{1}{c |}{\specialcell{\bf GQ-Adv-Phys}} & \multicolumn{1}{c |}{\specialcell{\bf GQ-Adv-FC}} & \multicolumn{1}{c |}{\specialcell{\bf GQ-Adv-LowU}} & \multicolumn{1}{c}{\specialcell{\bf GQ-Adv-HighU}} \\
       \hline & & & & & & & & & & & & \\
       {\bf Success Rate ($\%$)} & 58$\pm$11 & 70$\pm$10 & 75$\pm$9 & 80$\pm$9 & 95$\pm$5 & {\bf 93$\pm$6} & 85$\pm$8 & 83$\pm$8 & 80$\pm$9 & 83$\pm$8 & 78$\pm$9 & 86$\pm$8\\
       {\bf Precision ($\%$)} & N/A & N/A & 100 & 100 & N/A & {\bf 94} & 90 & 91 & 80 & 89 & 90 & 92  \\
       {\bf Robust Grasp Rate ($\%$)} & N/A & N/A & 5 & 0 & N/A & 43 & 60 & 44 & 100 & 89 & 53 & 64  \\
       {\bf Planning Time (sec)} & N/A & 1.9 & 0.8 & 0.9 & 2.6 & {\bf 0.8} & 0.9 & 0.8 & 0.8 & 0.7 & 0.8 & 0.9 \\
        \end{tabular}}
        \caption{Performance of grasp planning methods on the Train dataset with 95$\%$ confidence intervals for the success rate. Each method was tested for 80 trials (10 trials per object). Details on the methods used for comparison can be found in \secref{baselines}. We see that REG (point cloud registration) has the highest success rate at 95$\%$ but the GQ-L-Adv performs comparably at 93$\%$ and is 3$\times$ faster. Performance of the GQ-CNN drops to 80$\%$ when trained on the Adv-Phys dataset (GQ-Adv-Phys), likely due to the small number of training examples, and drops to 78$\%$ when no noise is added to the images during training (GQ-Adv-LowU).  }
		\tablabel{phys-val}
\end{table*}

\subsection{Grasp Planning Methods Used for Comparison}
\seclabel{baselines}
We compared a number of grasp planning methods on simulated and real data.
We tuned the parameters of each method based on synthetic classification performance and physical performance on the training objects.
All methods other than point cloud registration used the antipodal grasp sampling method described in \secref{planning} with the same set of parameters to generate candidate grasps, and each planner executes the highest-ranked grasp according to the method.
Additional details on the methods and their parameters can be found in the supplemental file.

{\bf Image-based Grasp Quality Metrics (IGQ).}
We sampled a set of force closure grasp candidates by finding antipodal points on the object boundary~\cite{chen1993finding} using edge detection and ranked grasps by the distance from the center of the jaws to the centroid of the object segmentation mask.
We set the gripper depth using a fixed offset from the depth of the grasp center pixel.

{\bf Point-Cloud Registration (REG).}
We also compared with grasp planning based on point cloud registration, a state-of-the-art method for using precomputed grasps~\cite{goldfeder2011data, hernandez2016team}.
We first coarsely estimated the object instance and pose based on the top 3 most similar synthetic images from Dex-Net 2.0, where similarity is measured as distance between AlexNet conv5 features~\cite{goldfeder2011data, mahler2016dexnet}.
After coarse matching, we finetuned the pose of the object in the table plane using Iterated Closest Point~\cite{gupta2015aligning, kehoe2013cloud} with a point-to-plane cost.
Finally, we retrieved the most robust gripper pose from Dex-Net 2.0 for the estimated object.
The system had a median translational error of 4.5$mm$ a median rotational error of 3.5$^{\circ}$ in the table plane for known objects.

{\bf Alternative Machine Learning Models (ML).}
We also compared the performance of a Random Forest with 200 trees of depth up to 10 (ML-RF) motivated by the results of~\cite{seitalarge} and a Support Vector Machine with the RBF kernel and a regularization parameter of 1 (ML-SVM) motivated by the results of~\cite{bohg2010learning, saxena2008robotic, ten2015using}.
For the RF we used the raw transformed images and gripper depths normalized by the mean and standard deviation across all pixels as features.
For the SVM we used a Histogram of Oriented Gradients (HOG)~\cite{dalal2005histograms} feature representation.
Both methods were trained using scikit-learn on the Adv-Synth dataset.

{\bf Grasp Quality CNNs (GQ).}
We trained the GQ-CNN (abbrev. GQ) using the thresholded robust epsilon metric with $\delta=0.002$~\cite{kappler2015leveraging} for 5 epochs on Dex-Net-Large (all of Dex-Net 2.0) using Gaussian process image noise with standard deviation $\sigma = 0.005$.
We used TensorFlow~\cite{abadi2016tensorflow} with a batch size of $128$, a momentum term of $0.9$, and an exponentially decaying learning rate with step size $0.95$.
Training took approximately 48 hours on an NVIDIA GeForce 1080.
The first layer of $7\times7$ convolution filters are shown in the right panel of \figref{architecture}, and suggest that the network learned  finegrained vertical edge detectors and coarse oriented gradients.
We hypothesize that vertical filters help to detect antipodal contact normals and the coarse oriented gradients estimate collisions.

To benchmark the architecture outside of our datasets, we trained on the Cornell Grasping Dataset~\cite{lenz2015deep} (containing 8,019 examples) and achieved a $93.0\%$ recognition rate using grayscale images and an $80-20$ imagewise training-validation split compared to $93.7\%$ on RGB-D images in the original paper.
We also trained several variants to evaluate sensitivity to several parameters:

{\it Dataset Size.}
We trained a GQ-CNN on Dex-Net-Small for 15 epochs (GQ-S). 

{\it Amount of Pretraining}
We trained three GQ-CNNs on the synthetic dataset of adversarial training objects (Adv-Synth) to study the effect of pretraining with Dex-Net for a new, known set of objects.
The model GQ-Adv was trained on only Adv-Synth for 25 epochs. 
The models GQ-L-Adv and GQ-S-Adv were initialized with the weights of GQ and GQ-S, respectively, and finetuned for 5 epochs on Adv-Synth.

{\it Success Metric.}
We trained a GQ-CNN using probability of force closure thresholded at 25$\%$ (GQ-Adv-FC), which is a robust version of the antipodality metric of~\cite{gualtieri2016high, ten2015using}, and and labels for 400 random grasp attempts on the Train objects using a physical robot~\cite{levine2016learning, pinto2016supersizing} (GQ-Adv-Phys).

{\it Noise Levels.}
We trained a GQ-CNN with zero noise $\sigma = 0$ (GQ-Adv-LowU) and high noise with $\sigma = 0.01$ (GQ-Adv-HighU).

\subsection{Classification of Synthetic Data}
The GQ-CNN trained on all of Dex-Net 2.0 had an accuracy of 85.7$\%$ on a held out validation set of approximately 1.3 million datapoints.
Due to the memory and time requirements of training SVMs, we compared synthetic classification performance across methods on the smaller Adv-Synth dataset.
\figref{roc} shows the receiver operating characteristic curve comparing the performance of GQ-L-Adv, GQ-S-Adv, GQ-Adv, ML-SVM, and ML-RF on a held-out validation set and \tabref{synth-val} details the classification accuracy for the various methods.
The GQ-CNNs outperformed ML-RF and ML-SVM, achieving near-perfect validation accuracy.

\subsection{Performance Comparison on Known Objects}
We evaluated the performance of the grasp planning methods on known objects from Train.
Each grasp planner had 80 trials (10 per object).
The left half of \tabref{phys-val} compares the performance with other grasp planning methods and the right half compares the performance of the GQ-CNN varations.
We found that GQ planned grasps 3$\times$ faster than REG and achieved a high 93$\%$ success rate and 94$\%$ precision.
The results also suggest that training on the full Dex-Net 2.0 dataset was necessary to achieve higher than $90\%$ success.

\subsection{Performance Comparison on Novel Objects}
We also compared the performance of the methods on the ten novel test objects from Test to evaluate generalization to novel objects.
Each method was run for 50 trials (5 per object). 
The parameters of each method were set based on Train object performance without knowledge of the performance on Test.
\tabref{phys-novel} details the results.
GQ performed best with an 80$\%$ success rate and $100\%$ precision (zero false positives over $29$ grasps classified as robust).

\begin{table}[t]
\centering
\resizebox{\textwidth}{!}{%
        \begin{tabular}{r || R{1.0cm} | R{1.0cm} | R{1.0cm} | R{1.0cm} | R{1.0cm} | R{1.0cm} | R{1.0cm} }
       \multicolumn{1}{c ||}{\specialcell{}}&  \multicolumn{1}{c |}{\specialcell{\bf IGQ}} & \multicolumn{1}{c |}{\specialcell{\bf REG}} & \multicolumn{1}{c |}{\specialcell{\bf GQ-Adv-Phys}} & \multicolumn{1}{c |}{\specialcell{\bf GQ-Adv}} & \multicolumn{1}{c |}{\specialcell{\bf GQ-S}} & \multicolumn{1}{c |}{\specialcell{\bf GQ}} \\
       \hline & & & & & & \\
       {\bf Success Rate ($\%$)} & 60$\pm$13 & 52$\pm$14 & 68$\pm$13 & 74$\pm$12 & 72$\pm$12 & {\bf 80$\pm$11}\\
       {\bf Precision ($\%$)} & N/A & N/A & 68 & 87 & 92 & {\bf 100} \\
       {\bf Robust Grasp Rate ($\%$)} & N/A & N/A & 100 & 30 & 48 & \bf 58 \\
       {\bf Planning Time (sec)} & 1.8 & 3.4 & 0.7 & 0.7 & 0.8 & 0.8 \\
        \end{tabular}}
        \caption{Performance of grasp planning methods on our grasping benchmark with the test dataset of 10 household objects with 95$\%$ confidence intervals for the success rate. Each method was tested for 50 trials, and details on the methods used for comparison can be found in \secref{baselines}. GQ performs best in terms of success rate and precision, with 100$\%$ precision (zero false positives among $29$ positive classifications). Performance decreases with smaller training datasets, but the GQ-CNN methods outperform the image-based grasp quality metrics (IGQ) and point cloud registration (REG).
        }
		\tablabel{phys-novel}
\end{table}

\begin{figure}[t!]
\centering
\includegraphics[scale=0.21]{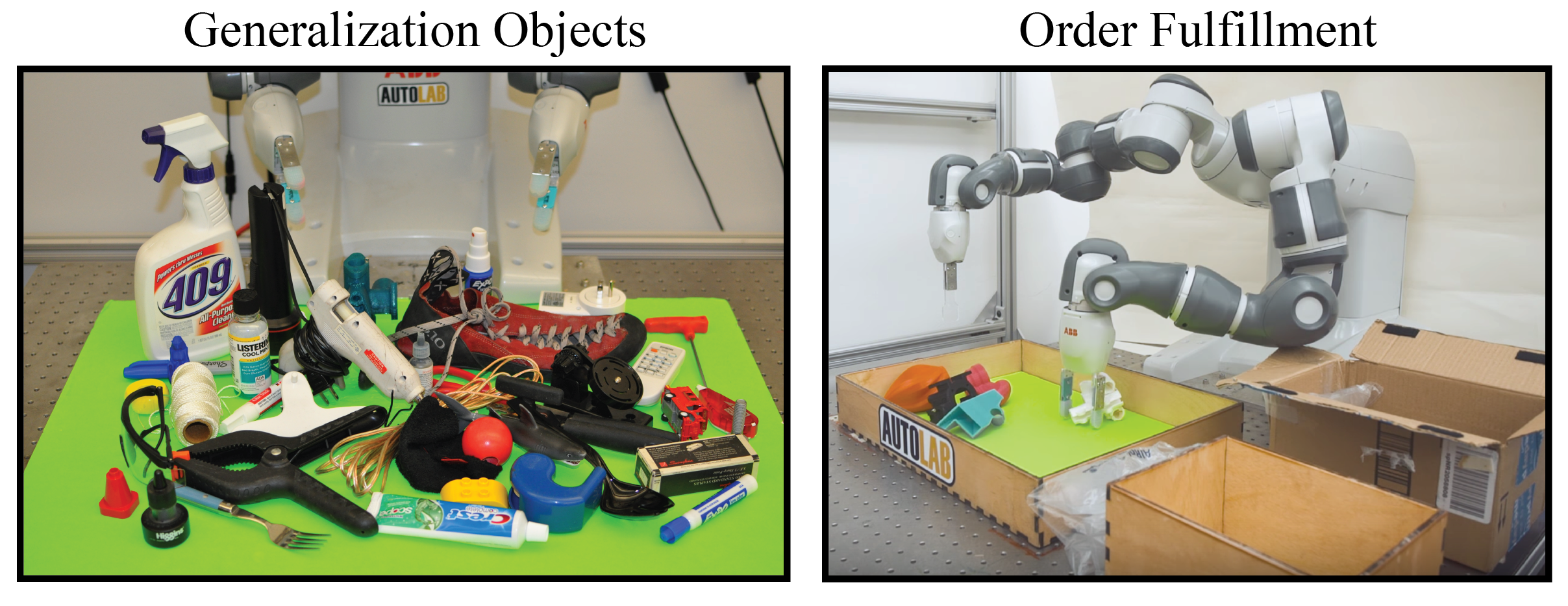}
\caption{(Left) The test set of 40 household objects used for evaluating the generalization performance of the Dex-Net 2.0 grasp planner. The dataset contains rigid, articulated, and deformable objects. (Right) The experimental setup for order fulfillment with the ABB YuMi. The goal is to grasp and transport three target objects to a shipping container (box on right). }
\figlabel{extensions}
\end{figure}

\subsection{Generalization Ability of the Dex-Net 2.0 Grasp Planner}
We evaluated the generalization performance of GQ in 100 grasping trials on the 40 object test set illustrated in \figref{extensions}, which contains articulated (e.g. can opener) and deformable (e.g. washcloth) objects.
We used the cross entropy method (CEM)~\cite{levine2016learning}, which iteratively samples a set of candidate grasps and re-fits the candidate grasp distribution to the grasps with the highest predicted robustness, in order to find better maxima of the robust grasping policy.
More details can be found in the supplemental file.
The CEM-augmented Dex-Net 2.0 grasp planner achieved 94$\%$ success and 99$\%$ precision (68 successes out of 69 grasps classified as robust), and it took an average of 2.5$s$ to plan grasps.

\subsection{Application: Order Fulfillment}
To demonstrate the modularity of the Dex-Net 2.0 grasp planner, we used it in an order fulfillment application with the ABB YuMi.
The goal was to grasp and transport a set of three target objects to a shipping box in the presence of three distractor objects when starting with the objects in a pile on a planar worksurface, illustrated in \figref{extensions}.
Since the Dex-Net 2.0 grasp planner assumes singulated objects, the YuMi first separated the objects using a policy learned from human demonstrations mapping binary images to push locations~\cite{laskey2016comparing}.
When the robot detected an object with sufficient clearance from the pile, it identified the object based on color and used GQ-L-Adv to plan a robust grasp.
The robot then transported the object to either the shipping box or a reject box, depending on whether or not the object was a distractor.
The system successfully placed the correct objects in the box on 4 out of 5 attempts and was successful in grasping on 93$\%$ of 27 total attempts.

\subsection{Failure Modes}
\figref{failures} displays some common failures of the GQ-CNN grasp planner.
One failure mode occured when the RGB-D sensor failed to measure thin parts of the object geometry, making these regions seem accessible.
A second type of failure occured due to collisions with the object.
It appears that the network was not able to fully distinguish collision-free grasps in narrow parts of the object geometry.
This suggests that performance could be improved with more accurate depth sensing and using analytic methods to prune grasps in collsion.

\begin{figure}[t!]
\centering
\includegraphics[scale=0.2]{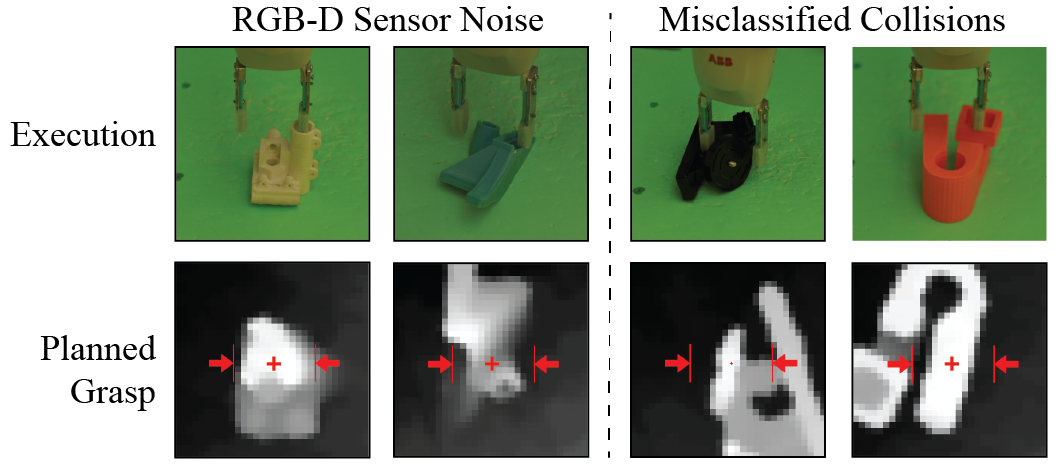}
\caption{Four examples of failed grasps planned using the GQ-CNN from Dex-Net 2.0. The most common failure modes were related to: (left) missing sensor data for an important part of the object geometry, such as thin parts of the object surface, and (right) collisions with the object that are misclassified as robust.}
\figlabel{failures}
\end{figure}

\section{Discussion and Future Work}
\seclabel{discussion}
We developed a Grasp Quality Convolutional Neural Network (GQ-CNN) architecture that predicts grasp robustness from a point cloud and trained it on Dex-Net 2.0, a dataset containing 6.7 million point clouds, parallel-jaw grasps, and robust grasp metrics.
In over 1,000 physical evaluations, we found that the Dex-Net 2.0 grasp planner is as reliable and 3$\times$ faster a method based on point cloud registration, and had $99\%$ precision on a test set of 40 novel objects.

In future work, our goal is to approach 100$\%$ success on known objects by using active learning to adaptively acquire grasps using a policy initialized with a GQ-CNN.
Additionally, we plan to exend the method to grasp objects in clutter~\cite{gualtieri2016high, levine2016learning} by using simulated piles of rigid objects from Dex-Net and by augmenting the grasping policy with an option to push and separate objects when no robust grasp is available.
We also intend to extend the method to use point clouds from multiple viewpoints and in grasping tasks with sequential structure, such as regrasping for assembly.
Furthermore, we plan to release a subset of our code, dataset, and the trained GQ-CNN weights to facilitate further research and comparisons.

\section*{Acknowledgments}
{\tiny
This research was performed at the AUTOLAB at UC Berkeley in affiliation with the Berkeley AI Research (BAIR) Lab, the Real-Time Intelligent Secure Execution (RISE) Lab, and the CITRIS ”People and Robots” (CPAR) Initiative.
The authors were supported in part by the U.S. National Science Foundation under NRI Award IIS-1227536: Multilateral Manipulation by Human-Robot Collaborative Systems, the Department of Defense (DoD) through the National Defense Science \& Engineering Graduate Fellowship (NDSEG) Program, the Berkeley Deep Drive (BDD) Program, and by donations from Siemens, Google, Cisco, Autodesk, IBM, Amazon Robotics, and Toyota Robotics Institute.
Any opinions, findings, and conclusions or recommendations expressed in this material are those of the author(s) and do not necessarily reflect the views of the Sponsors.
We thank our colleagues who provided helpful feedback, code, and suggestions, in particular Pieter Abbeel, Ruzena Bajcsy, Brenton Chu, Roy Fox, David Gealy, Ed Johns, Sanjay Krishnan, Animesh Garg, Sergey Levine, Pusong Li, Matt Matl, Stephen McKinley, Andrew Reardon, Vishal Satish, Sammy Staszak, and Nan Tian.
}
\appendices
\section{Parameters of Graphical Model}
\label{graphical-details}

Our graphical model is illustrated in \figref{graphical-model} and models $p(S, \bg, \bx, \by)$ as the product of a state distribution $p(\bx)$, an observation model $p(\by | \bx)$, a grasp candidate model $p(\bg | \bx)$, and a grasp success model $p(S | \bg, \bx)$.

We model the state distribution as $p(\bx) = p(\gamma) p(\mO) p(T_{o} | \mO)  p(T_{c})$.
We model $p(\gamma)$ as a Gaussian distribution $\mN(0.5, 0.1)$ truncated to $[0,1]$.
We model $p(\mO)$ as a discrete uniform distribution over 3D objects in a given dataset.
We model $p(T_{o} | \mO) = p(T_{o} | T_{s}) p(T_{s} | \mO) $, where is $p(T_{s} | \mO)$ is a discrete uniform distribution over object stable poses and $p(T_{o} | T_{s})$ is uniform distribution over 2D poses: $\mU([-0.1,0.1]\times[-0.1,0.1]\times[0,2\pi))$.
We compute stable poses using the quasi-static algorithm given by Goldberg et al.~\cite{goldberg1999part}.
We model $p(T_{c})$ as a uniform distribution on spherical coordinates $r, \theta, \phi \sim \mU([0.65, 0.75]\times[0,2\pi)\times[0.05\pi, 0.1\pi])$, where the camera optical axis always intersects the center of the table.

Our distribution over grasps is a uniform distribution over pairs of antipodal points on the object surface that are parallel to the table plane.
We sample from this distribution for a fixed coefficient of friction $\mu = 0.6$ and reject samples outside the friction cone or non-parallel to the surface.

We model images as $\by = \alpha * \hat{\by} + \epsilon$ where $\hat{\by}$ is a rendered depth image created using OSMesa offscreen rendering.
We model $\alpha$ as a Gamma random variable with shape$=1000.0$ and scale=$0.001$.
We model $\epsilon$ as Gaussian Process noise drawn with measurement noise $\sigma=0.005$ and kernel bandwidth $\ell = \sqrt{2}px$.

We compute grasp robustness metrics using the graphical model and noise parameters of~\cite{seitalarge}.

\section{Grasp Sampling Methods}
\label{grasp-sampling}
Our goal is to learn policy parameters $\theta$ that maximize the success rate of planned grasps over a distribution on point clouds that can be generated from a set of possible objects $\mathcal{D}$:
\begin{align}
	\theta^* = \myargmax{\theta \in \Theta} \mathbb{E}_{p(\by \mid \mathcal{D})} \left[Q(\pi_{\theta}, \by)\right]. \label{eq:obj}
\end{align}

We propose to approach the objective of equation~\ref{eq:obj} by using a greedy grasping policy $\pi_{\theta}(\by) = \text{argmax}_{\bu \in \mC} Q_{\theta}(\bu, \by)$ with respect to a GQ-CNN robustness function $Q_{\theta}(\bu, \by)$, where $\mC$ specifies constraints on candidate grasps such as kinematic feasibility.
We explore two implementations of the robust grasping policy: (1) sampling a large, fixed set of antipodal grasps and choosing the most robust one and (2) optimizing for the most robust grasp using derivative free optimization.

\subsection{Antipodal Grasp Sampling}
The antipodal grasp sampling method used in the paper is designed to sample antipodal grasps specified as a planar pose, angle, and height with respect to a table.
The algorithm is detailed in Algorithm~\ref{alg:sampling}.
We first threshold the depth image to find areas of high gradient.
Then, we use rejection sampling over pairs of pixels to generate a set of candidate antipodal grasps, incrementally increasing the friction coefficient until a desired number of grasps is reached in case the desired number cannot be achieved with a smaller friction coefficient.
We convert antipodal grasps in image space to 3D by assigning discretizing the gripper height between the height of the grasp center pixel relative and the height of the table surface itself.

This grasp sampling method is used for all image based grasp planners in the paper.
We used $M = 1000$, $K$ set to the intrinsics of a Primesense Carmine 1.08, $T_{c}$ determined by chessboard registration, $g = 0.0025m$, $\mu_{\ell} = 0.4$, $\delta_{\mu} = 0.2$, $N = 1000$, and $\delta_h = 0.01m$.

\begin{algorithm}
{\small
    \SetAlgoLined
    {\bf Input:} Depth image $\by$, Number of grasps $M$,  Camera Intrinsics Matrix $K$, Camera pose $T_{c}$, Depth gradient threshold $g$, Min friction coef $\mu_{\ell}$, Friction coef increment $\delta_{\mu}$, Max samples per friction coef $N$, Gripper height resolution $\delta_{h}$ \\
    \KwResult{$\mG$, set of candidate grasps}
    
    \tcp{Compute depth edges}
	$G_x = \nabla_x \by, G_y = \nabla_y \by$\;
	$\mE = \{\bu \in \bR^2 : G_x(\bu)^2 + G_y(\bu)^2 > g\}$\;    
	
	\tcp{Find antipodal pairs}
	$\mG = \{\}$, $i=0$, $j=0$\;
	\While{$|\mG| < M$ and $\mu <= 1.0$}{
		$\bu, \bv = $UniformRandom($\mE, 2$)\;
		\If{Antipodal($\bu, \bv, \mu$)}{
			\tcp{Compute piont in world coordinates}
			$\bc = 0.5 * (\bu + \bv)$\;
			$\bp_c = $Deproject($K, \by, \bc$)\;
			$\bp = T_{c} * \bp_c$\;
			$h = \bp.z$\;
			
			\tcp{Add all heights}
			\While{$h > 0$}{
				$\mG = \mG \cup \{\bg(\bu, \bv,  h)\}$\;
				$h = h - \delta_h$\;
			}
		}
		$i = i+1$, $j = j+1$\;
		
		\tcp{Update friction coef}
		\If{$j >= N$}{
			$\mu = \mu + \delta_{\mu}$\;
			$j = 0$\;
		}
    }
    return $\mG$\;
    
    \caption{Antipodal Grasp Sampling from a Depth Image}
    \label{alg:sampling}
}
\end{algorithm}

\subsection{Derivative Free Optimization}
One problem with choosing a grasp from a fixed set of candidates is that the set of candidates may all have a low probability of success. 
This can be difficult when an object can only be grasped in a small set of precise configurations, such as the example in \figref{cem}.
Some of these failures can be seen in the right panel of the failure modes figure in the original paper.

In our second generalization study we addressed this problem using the cross entropy method (CEM)~\cite{levine2016learning, rubinstein2013fast}, a form of derivative-free optimization, to optimize for the most robust grasp by iteratively resampling grasps from a learned distribution over robust grasps and updating the distribution.
The method, illustrated in Algorithm~\ref{alg:cem}, models the distribution on promising grasps using a Gaussian Mixture Model (GMM) and seeds the initial set of grasps with antipodal point pairs using Algorithm~\ref{alg:sampling} with no iterative friction coefficient updates. 
The algorithm takes as input the number of CEM iterations $m$, the number of initial grasps to sample $n$, the number of grasps to resample from the model $c$, the number of GMM mixure components $k$, a friction coefficient $\mu$, and elite percentage $\gamma$, and the GQ-CNN $Q_{\theta}$, and returns an estimate of the most robust grasp $\bu$.
In our generalization experiment we used $m=3$, $n=100$, $c=50$, $\mu=0.8$, $k=3$, and $\gamma=25\%$.
The qualitative performance of our method on several examples from our experiments is illustrated in \figref{cem-examples}.

\begin{figure}[t!]
\centering
\includegraphics[scale=0.2]{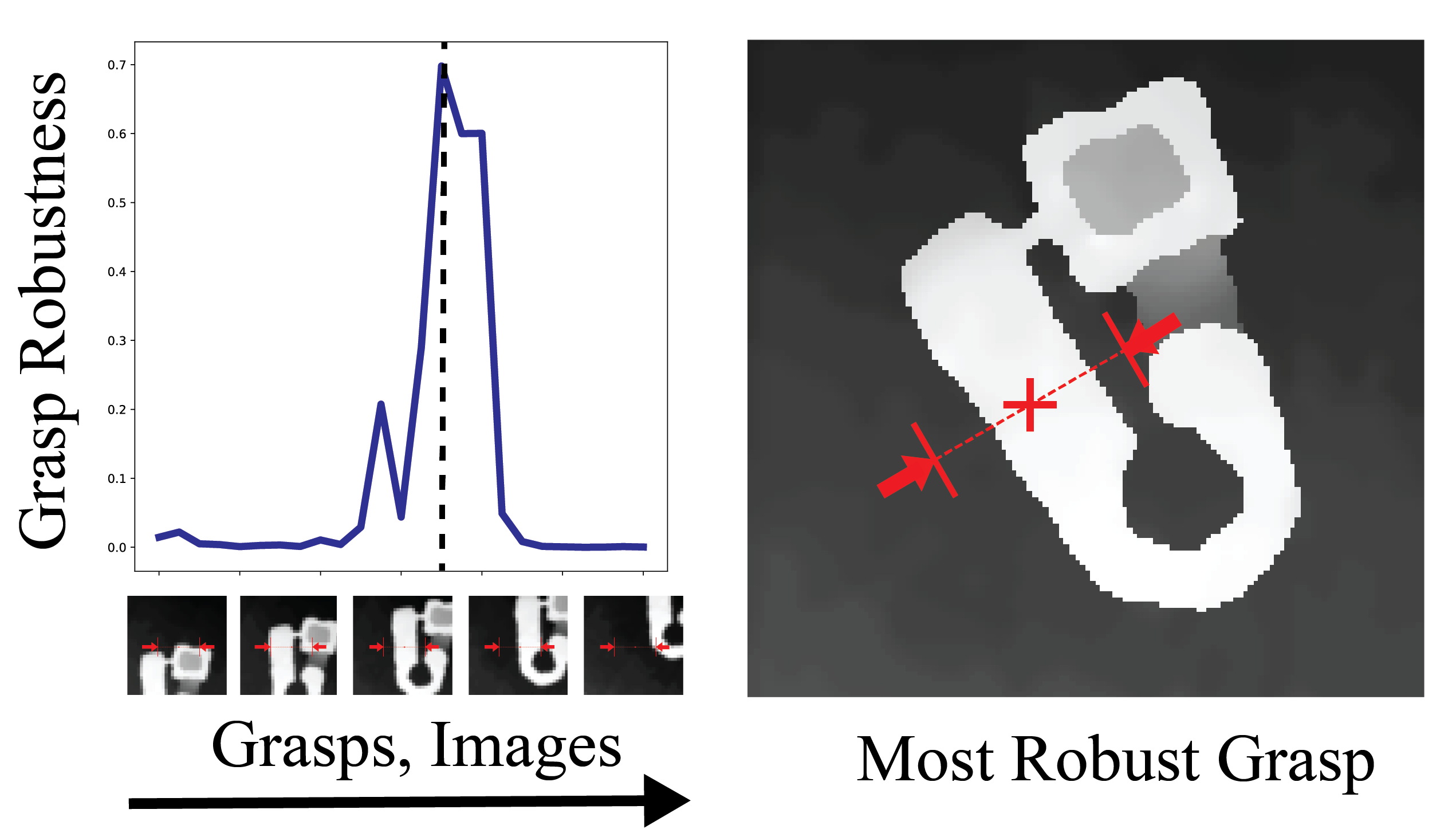}
\caption{(Left) Grasp robustness predicted by a Grasp Quality Convolutional Neural Network (GQ-CNN) trained with Dex-Net 2.0 over the space of depth images and grasps for a single point cloud collected with a Primesense Carmine. As the center of the gripper moves from the top to the bottom of the image the GQ-CNN prediction stays near zero and spikes on the most robust grasp (right), for which the gripper fits into a small opening on the object surface. This suggests that the GQ-CNN has learned a detailed representation of the collision space between the object and gripper. Furthermore, the sharp spike suggests that it may be difficult to plan robust grasps by randomly sampling grasps in image space. We consider planning the most robust grasp using the cross-entropy method on the GQ-CNN response.
\vspace*{-20pt}}
\figlabel{cem}
\end{figure}

\begin{figure}[t!]
\centering
\includegraphics[scale=0.9]{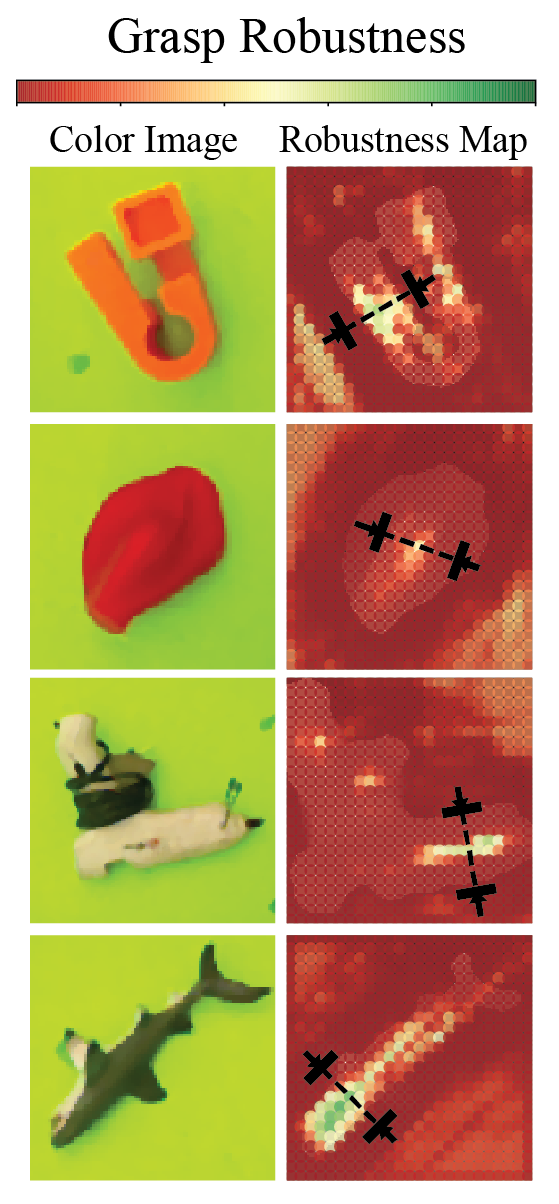}
\caption{Example input color images and maps of the grasp robust estimated by the GQ-CNN over grasp centers for a constant grasp axis angle in image space and height above the table, with the grasp planned by our CEM-based robust grasping policy shown in black. CEM is able to find precise robust grasping locations encoded by the GQ-CNN that are very close to the global maximum for the given grasp axis and height. The GQ-CNN also appears to assign non-zero robustness to several grasps that completely miss the object. This is likely because no such grasps are in the training set, and future work could augment the training dataset to avoid these grasps. }
\vspace*{-20pt}
\figlabel{cem-examples}
\end{figure}

\begin{algorithm}
{\small
    \SetAlgoLined
    {\bf Input:} Num rounds $m$, Num inital samples $n$, Num CEM samples $c$, Num GMM mixture $k$, Friction coef $\mu$, Elite percentage $\gamma$, Robustness function $Q_{\theta}$\\
    \KwResult{$\bu$, most robust grasp}
    $\mU \leftarrow $uniform set of $n$ antipodal grasps\;
    \For{i = 1, ..., m}{
    $\mE \leftarrow $top $\gamma-$percentile of grasps ranked by $Q_{\theta}$\;
    $M \leftarrow $GMM fit to $\mE$ with $k$ mixtures\;
    $G \leftarrow c$ iid samples from $M$\;
    }
    return $\myargmax{\bu \in \mU}{Q_{\theta}(\bu, \by)}$\;
    \caption{Robust Grasping Policy using the Cross Entropy Method on a Learned GQ-CNN}
    \label{alg:cem}
}
\end{algorithm}

\fontsize{8.8pt}{9.9pt} \selectfont
\bibliographystyle{plainnat}
\bibliography{bibliography}

\end{document}